\documentclass{cta-author}

{}
{}
{}

\def\reffig#1{Fig.~\ref{#1}}

\begin{document}

\supertitle{Special Issue on Computer Vision for the Creative Industry\\This paper is a postprint of a paper submitted to and accepted for publication in [journal] and is subject to Institution of Engineering and Technology Copyright. The copy of record is available at the IET Digital Library}

\title{Going beyond Free Viewpoint: Creating Animatable Volumetric Video of Human Performances}

\author{\au{Anna Hilsmann$^{1}$}, \au{Philipp Fechteler$^{1}$}, \au{Wieland Morgenstern$^{1}$}, \au{Wolfgang Paier$^{1}$}, \au{Ingo Feldmann$^{1}$}, \au{Oliver Schreer$^{1}$}, \au{Peter Eisert$^{12\corr}$}}

\address{\add{1}{Vision \& Imaging Technologies, Fraunhofer HHI, Berlin, Germany}
\add{2}{Visual Computing, Humboldt University, Berlin, Germany}
\email{peter.eisert@hhi.fraunhofer.de}}

\begin{abstract}
In this paper, we present an end-to-end pipeline for the creation of high-quality animatable volumetric video content of human performances.
Going beyond the application of free-viewpoint volumetric video, we allow re-animation and alteration of an actor's performance through (i) the enrichment of the captured data with semantics and animation properties and (ii) applying hybrid geometry- and video-based animation methods that allow a direct animation of the high-quality data itself instead of creating an animatable model that \textit{resembles} the captured data.
Semantic enrichment and geometric animation ability are achieved by establishing temporal consistency in the 3D data, followed by an automatic rigging of each frame using a parametric shape-adaptive full human body model. Our hybrid geometry- and video-based animation approaches combine the flexibility of classical CG animation with the realism of real captured data. For pose editing, we exploit the captured data as much as possible and kinematically deform the captured frames to fit a desired pose. Further, we treat the face differently from the body in a hybrid geometry- and video-based animation approach where coarse movements and poses are modeled in the geometry only, while very fine and subtle details in the face, often lacking in purely geometric methods, are captured in video-based textures. These are processed to be interactively combined to form new facial expressions. On top of that, we learn the appearance of regions that are challenging to synthesize, such as the teeth or the eyes, and fill in missing regions realistically in an autoencoder-based approach. This paper covers the full pipeline from capturing and producing high-quality video content, over the enrichment with semantics and deformation properties for re-animation and processing of the data for the final hybrid animation.

\end{abstract}

\maketitle

\section{Introduction}\label{sec_intro}
In recent years, with the advances in augmented and virtual reality, high-quality free viewpoint video has gained a lot of interest.
Today, volumetric studios \cite{web_microsoft,web_volucap} can create high-quality 3D video content and advanced virtual/augmented reality hardware \cite{web_oculus, web_vive} is able to create highly immersive viewing experiences.
However, high-quality immersiveness is usually limited to experience pre-recorded situations. Changing the recorded scene, e.g. the motion and performance of captured persons, and hence, interaction with high-quality volumetric video (VV) content is usually not possible. In this paper, we think immersiveness and interactivity a step further by making high quality captured human performances alterable and editable.

The vision of making captured content animatable has gained a lot of attention in the computer graphics community in recent years. The usual approach is to fit a generic body model to the captured volumetric data and use this fitted model that approximates the captured data for animation \cite{achenbach2017, feng2015}. Using statistical models, animations can even be created from monocular video \cite{weng19}.  In this paper, we follow a different approach and present a full end-to-end pipeline for the creation of alterable high-quality volumetric video content of human performances: instead of creating an animatable model that approximates the captured data, we propose to exploit the captured high-quality real-world data as much as possible as it contains all-natural deformations and characteristics. This is achieved by enriching the captured high-quality data with semantics and animation properties, allowing animation of the high-quality real-world data \textit{itself}.

For animation, we propose a novel hybrid example-based approach, that exploits the captured content as much as possible to create new animations and motion sequences in a photorealistic and natural way.
First, the semantic enrichment allows direct manipulation of the captured volumetric video frames by kinematic animation.
Further, we propose to subdivide the captured volumetric video into elementary subsequences to form new animation sequences through appropriate concatenation and interpolation together with kinematic manipulation of the individual frames.
As humans are very sensitive to viewing facial expressions, and it is very difficult to represent all fine and subtle details of facial movements in geometry, we propose a special hybrid geometry-and video-based approach for the creation of new facial animation sequences.
In our approach, the geometry accounts for low-resolution shape adaptation while fine details are captured by video-textures that function as appearance examples to be combined and interpolated to form new facial textures. Regions such as the eyes or the teeth are still difficult to model both in geometry as well as in texture due to e.g. (dis-)occlusion.
We address this by learning the appearance of these regions and realistically synthesize them in an autoencoder-based approach if they are missing in the manipulated texture.
With the proposed hybrid animation framework, we take advantage of the richness of the captured data that exhibits real deformations and appearances, producing animations with real shapes, kinematics and appearances.

This paper builds upon our previous works \cite{efr17,sfr19,fhe19,fkhe18,pkhe17,fphe16,pkhe15} and combines and extends them to build a full framework for the creation of animatable volumetric video (AVV).
The remainder of this paper is structured as follows. Sec.~\ref{sec_sota} gives an overview of related work before Sec.~\ref{sec_overview} presents an overview of the full pipeline.
Sec.~\ref{sec_volvideo} describes the capturing process in our volumetric studio and the creation of high-quality volumetric video data.
Sec.~\ref{sec_temporal} explains the temporal processing of the volumetric video stream in order to create temporally coherent mesh subsequences.
Sec.~\ref{sec_bodymodel} describes how we enrich the captured data with semantics and animation parameters by fitting a parametric human body model to it, before Sec.~\ref{sec_poseedit} covers the editing and pose-animation of the volumetric video content.
Finally, Sec.~\ref{sec_face} describes our proposed example based on hybrid facial animation.

\begin{figure*}[t]
\centering{\includegraphics{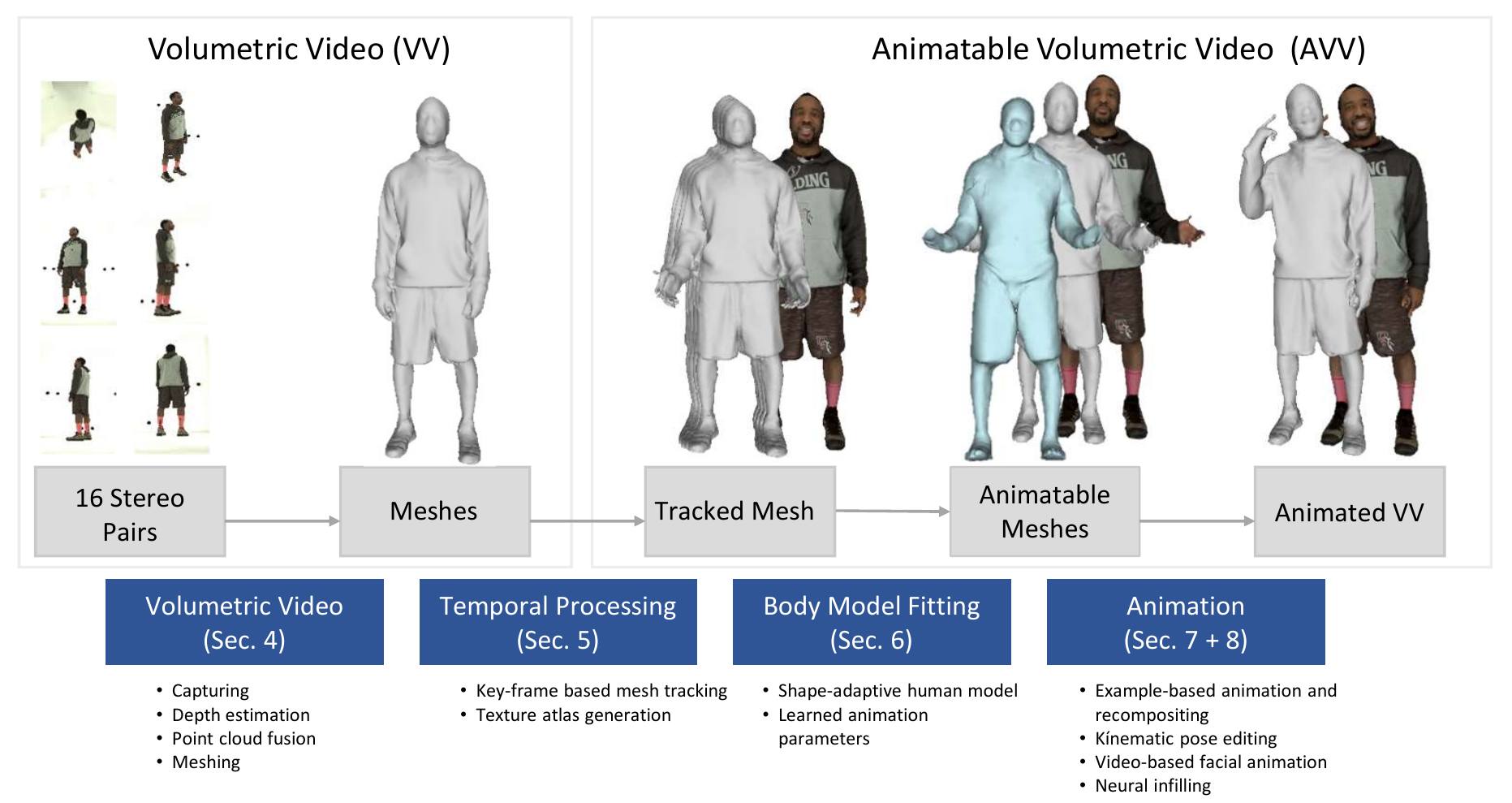}}
\caption{End-to-end pipeline for the creation of animatable volumetric video content.}\label{fig_overview}
\end{figure*}
\section{Related Work}\label{sec_sota}
We consider the problem of creating animatable volumetric video content, which is related to a number of different research topics. 
In the following, we will briefly review relevant work in the fields of volumetric video, human body modelling and animation and example-based animation techniques.\\

\textbf{Volumetric Video.} 
In recent years, a number of volumetric studios have been created \cite{web_microsoft,web_volucap,web_8i,web_uncorporeal,web_4dviews} that are able to produce high quality free viewpoint video that can be viewed in real-time from a continuous range of viewpoints chosen at any time in playback. 
Most studios focus on a capture volume that is viewed in 360 degrees from the outside. 
A large number of cameras are placed around the scene (e.g. in studios from 8i \cite{web_8i}, 4dviews \cite{web_4dviews}, uncorporeal \cite{web_uncorporeal}, or Volucap \cite{web_volucap}) providing input for photogrammetric
reconstruction of the actors, while Microsoft's Mixed reality Capture Studios \cite{web_microsoft} rely on active depth sensors for geometry acquisition. In order to separate the scene from the background, all studios are equipped with green screens for chroma keying. Only Volucap \cite{web_volucap} uses a bright backlit background to avoid green spilling effects in the texture.

In addition to these professional studios, also lighter and cheaper solutions have been proposed. For example, Casas et al.~\cite{cvch14} produce volumetric video data with only 8 cameras and use the captured data for examples based animation (see below), and Robertini et al.~\cite{robertini16} produce volumetric video captured in outdoor scenes. Alldieck et al.~\cite{alldieck2018video, alldieck2019} as well as Xu et al.~\cite{Xu2018} obtain accurate 3D body models and texture of arbitrary people from monocular video.

Usually, the content creation starts with the reconstruction of temporally inconsistent 3D reconstructions per frame, sometimes followed by surface matching techniques to produce temporally consistent meshes \cite{collet2015high}. The dynamic textured 3D reconstructions can then be inserted as assets in virtual environments and viewed there from arbitrary directions. 
However, animation, pose modifications or interaction with the content is not possible.\\

\textbf{Human Body Modelling and Animation.}
When animation of virtual humans is required, as it is the case for applications
like computer games, virtual reality or film, usually computer graphics models are used.
They allow for arbitrary animation, with body motion usually being controlled by an underlying skeleton while facial expressions are described by a set of blendshapes \cite{fwb18}.
The advantage of full control comes at the price of significant modelling effort and sometimes limited realism.

In this paper, we combine CG human body models with volumetric video, resulting in a hybrid representation that can be animated while preserving high realism from the captured data. 
This requires the body model to be adapted in shape and pose to the volumetric video performance.
Given a template model, shape and pose can be learned from the sequence of real 3D measurements \cite{SMPL2015,fhe19} in order to align it with the sequence.
Recent progress in deep learning also enables the reconstruction of highly accurate human body models
even from single RGB images \cite{ambt19}.
Similarly, Pavlakos et al.~\cite{pcgb19} estimates shape and pose of a template model from monocular video such that 
the human model exactly follows the performance in the sequence.
Haberman et al.~\cite{hxzm19} go one step further and enable real-time capture of humans including surface deformations due to clothes.
All these approaches provide sequences of animation parameters that can be applied to computer graphics models in order to replicate the actor's motion. 
The reproduction of very subtle details in motion and deformation, however, are beyond the scope of these approaches, but can be recovered using real captured data.  \\


\textbf{Example-based Animation Synthesis.}
Recently, more and more hybrid and example-based animation synthesis methods have been proposed that exploit captured data in order to obtain realistic appearances. 

One of the first example-based methods has been presented by \cite{bsc97} and \cite{sse00}, who synthesize novel video sequences of facial animations and other dynamic scenes by video resampling. 
Malleson et al.~\cite{mbwb15} present a method to continuously and seamlessly blend between multiple facial performances of an actor by exploiting complementary properties of audio and visual cues to automatically determine robust correspondences between takes, allowing a director to generate novel performances after filming. 
These methods yield 2D photorealistic synthetic video sequences, but are limited to replaying captured data. 
This restriction is overcome by Fyffe et al.~\cite{fjai14} and Serra et al.~\cite{scro18}, who use a motion graph in order to interpolate between different 3D facial expressions captured and stored in a database.

For full body poses, Xu et al.~\cite{xls11} introduced a flexible approach to synthesize new sequences for captured data by matching the pose of a query motion to a dataset of captured poses and warping the retrieved images to query pose and viewpoint. 
Combining image-based rendering and kinematic animation, photo-realistic animation of clothing has been demonstrated from a set of 2D images augmented with 3D shape information in \cite{hfe13}. 
Similarly, Thies et. al.~\cite{thies16} perform facial reenactment by infilling and deformation transfer of the mouth region from a database (input sequence) to fit the target expression and Paier et al.~\cite{pkhe17} combine blendshape-based animation with recomposing video-textures for the generation of facial animation. 

Character animation by resampling of 4D volumetric video has been investigated by \cite{smh05,hhs09}, yielding high visual quality. However, these methods are limited to replaying segments of the captured motions. 
In \cite{sgat10}, Stoll et al.~combine skeleton-based CG models with captured surface data to represent details of apparels on top of the body.
Casas et al.~\cite{cvch14} combined  concatenation of captured 3D sequences with view dependent texturing for real-time interactive animation. 
Similarly, Volino et al.~\cite{vhh15} presented a parametric motion graph-based character animation for web applications. Only recently, Boukhayma and Boyer \cite{bb15,bb19} proposed an animation synthesis structure for the recomposition of textured 4D video capture, accounting for geometry and appearance. 
They propose a graph structure that enables interpolation and traversal between precaptured 4D video sequences.
Finally, Regateiro et al.~\cite{rvh18} present a skeleton-driven surface registration approach to generate temporally consistent meshes from volumetric video of human subjects in order to facilitate intuitive editing and animation of volumetric video.\\

\textbf{Neural Animation.} Purely data driven methods have recently gained significant importance due to the progress in deep learning and the possibility to synthesize images and video. 
Chan et al.~\cite{cgze19}, for example, use 2D skeleton data to transfer body motion from one person to another and synthesize new videos with a generative adversarial network. 
The skeleton motion data can also be estimated from video by neural networks \cite{mssr17}.
Liu et al.~\cite{lxzk19} extend that approach and use a full template model as an intermediate representation that is enhanced by the GAN.
In \cite{Tripathy2019}, Tripathy et al.~present a GAN based approach for controllable facial reenactment using an interpretable facial attribute vector that consists of head pose (roll, pitch, yaw) action unit activations.
Another approach for facial reenactment (face swapping) is presented by Nirkin et al.~\cite{nirkin2019}, where they propose a RNN-based method that automatically adjusts for pose and expression variations.
Similar techniques can also be used for synthesizing facial video as shown, e.g. in \cite{kgtx18}.
While these approaches show astonishing and highly realistic results, viewpoints are mostly restricted to the original camera position when capturing training data.\\

In this paper, we combine these different techniques and present a full pipeline for the capturing and animation of volumetric video data. We exploit example-based animation together with CG models for semantic control and deep learning for facial synthesis. In this cascadic hybrid animation approach, original data is used as much as possible in order to achieve high realism, while approximate CG models give us full control for modification and animation.


\section{System Overview}\label{sec_overview}


An overview of our end-to-end pipeline is presented in \reffig{fig_overview} and in the accompanying video.

In the following, we will step through the different components in our processing pipeline. Starting from the capturing of volumetric video data (Sec.~\ref{sec_volvideo}) and the processing into a temporally coherent sub-sequences of 3D meshes (Sec.~\ref{sec_temporal}), the next step is to make the captured volumetric video stream animatable and alterable. We address this by enriching the captured data with semantics and animation properties by fitting a parametric kinematic human body model to it (Sec.~\ref{sec_bodymodel}). The output of this step is a volumetric video stream with an attached parametric body model. Note, that the proposed pipeline is highly modular. For example, for the capturing and data processing steps our approach focuses on high quality professional capturing, which requires a sophisticated setup in a controlled setting. Generally, for these steps, also alternative methods for the creation of volumetric video as described in Sec.~\ref{sec_sota} can be used, considering the trade-off between data quality and captured details and the complexity of the capturing setup.

Through the enrichment of the volumetric video data with kinematic information, we can directly manipulate the captured volumetric frames themselves. Following, we discuss hybrid animation frameworks for the modification of the captured volumetric video data in Sec.~\ref{sec_poseedit}. Further, we propose to subdivide the captured data into elementary sub-sequences that can be concatenated and interpolated to form new animation sequences in combination with kinematic animation. Finally, as humans are especially sensitive to viewing human faces and facial expressions, we treat the human face separately from the body and present a hybrid geometry- and video-based approach for the creation of novel facial animations (Sec.~\ref{sec_face}).

\section{Volumetric Video}\label{sec_volvideo}

\begin{figure}[htb]
\centering{\includegraphics[width=\linewidth]{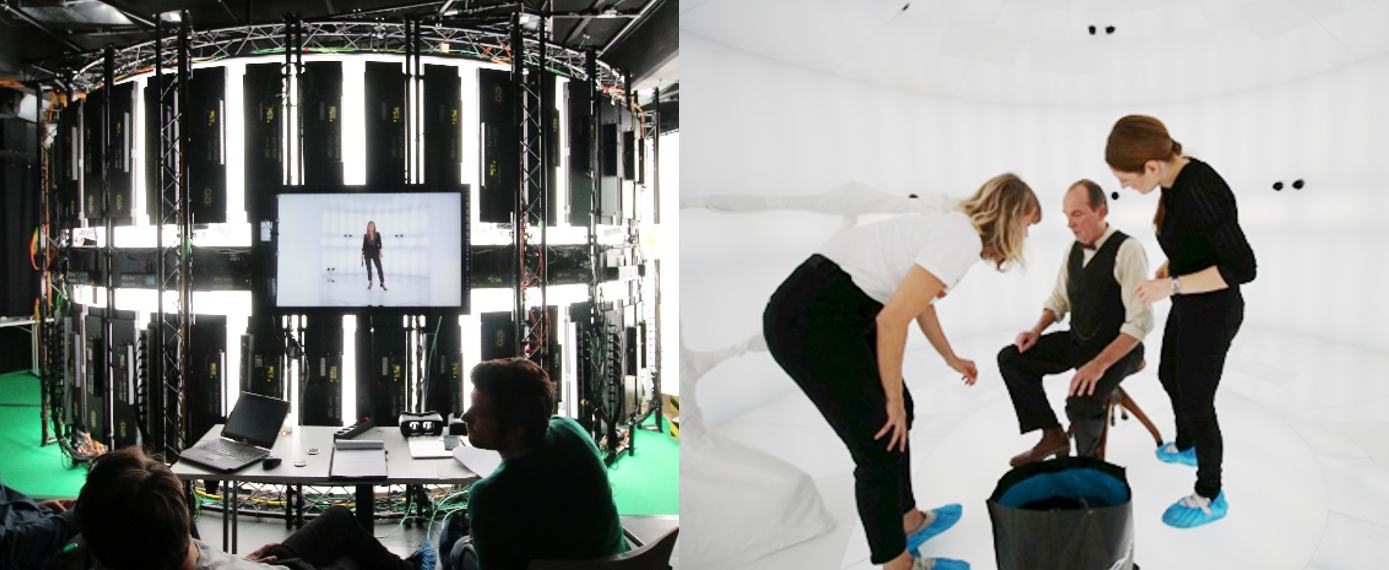}}
\caption{Volumetric studio for performance capturing.\label{fig_volvideo1}}
\end{figure}

The first step in our processing pipeline is the capturing of an actor's performance and the computation of a temporal sequence of 3D meshes.
For the volumetric capture, we have set up a studio with 32 cameras and 120 LED panels for full 360 degree acquisition as shown in \reffig{fig_volvideo1}.
The cameras are equipped with high-quality sensors offering 20 MPixel resolution at 30 frames per second, which enables pure image-based 3D reconstruction with high texture resolution.
They are grouped into 16 stereo pairs, positioned in 3 rings of different height.
From each stereo system, depth maps are computed, which are then fused into complete 3D models \cite{efr17, sfk19}.
Shape estimation is supported by a visual hull formed from segmentation masks created by automatic keying.
In contrast to other studios, we avoid green screen but segment the foreground against a bright diffuse background.
In addition, 120 programmable light panels offer homogeneous and diffuse lighting but also the creation of arbitrary lighting situations similar to a light stage.

\subsection{Stereo-based Depth Estimation}

The processing pipeline towards a sequence of 3D meshes representing the dynamic geometry of the actors starts with the computation of depth maps for each pair of cameras.
For depth estimation from the 5k by 4k images, a recursive patch-based method is used \cite{wfskw16,wfse13}.
Small patches around a pixel in a reference frame are swept through depth with different patch orientations and evaluated by normalized cross correlation in the second view, exploiting epipolar constraints obtained from calibration.
The large space of possible patch combinations (depth / orientation) is reduced by a multi-hypothesis method that propagates information from a spatial and temporal neighborhood.
Instead of testing all combinations, only depth/orientation hypotheses from other neighboring patches and from the previous frame are evaluated.
In addition, a new random candidate is considered similar as in \cite{bsfg09} and a flow-based depth/orientation refinement step is applied to the selected best hypothesis.
Since propagation of hypotheses is only performed from the previous step of iteration, no data dependencies exist, enabling massive parallel execution on multiple GPUs.
For each frame, a few steps of iteration are followed by left-right consistency checks of the stereo pair's two depth maps.
The full stereo-based depth estimation approach is detailed in \cite{wfskw16}.

\subsection{Point Cloud Fusion and Meshing}

\begin{figure}[htb]
\centering{\includegraphics[ height = 4.5cm ]{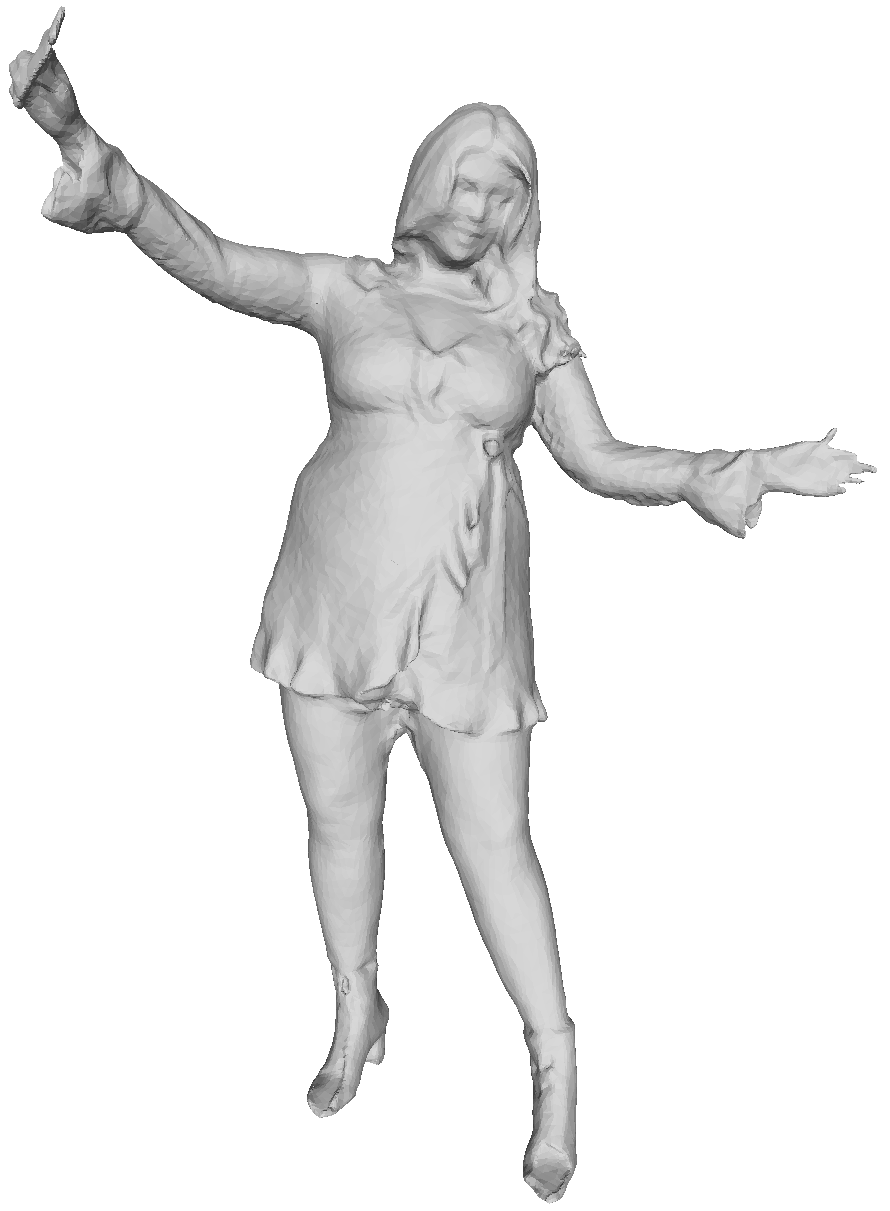}}
\centering{\includegraphics[ height = 4.2cm ]{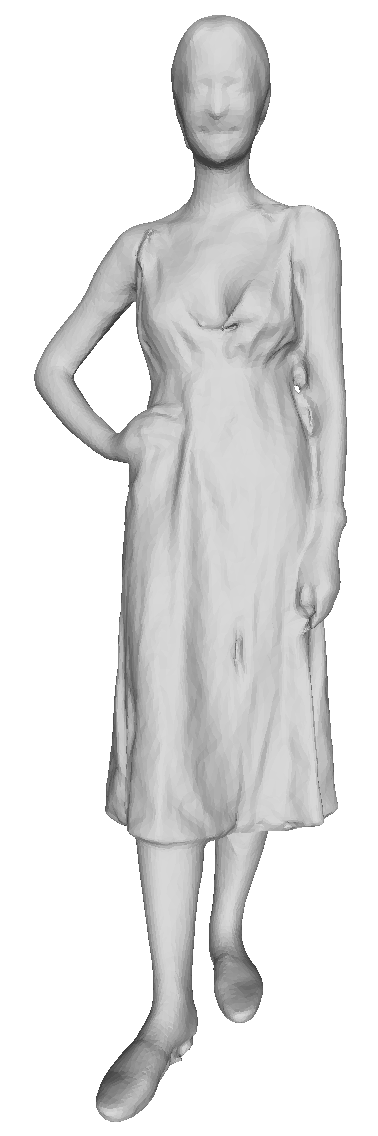}}
\centering{\includegraphics[ height = 4.2cm ]{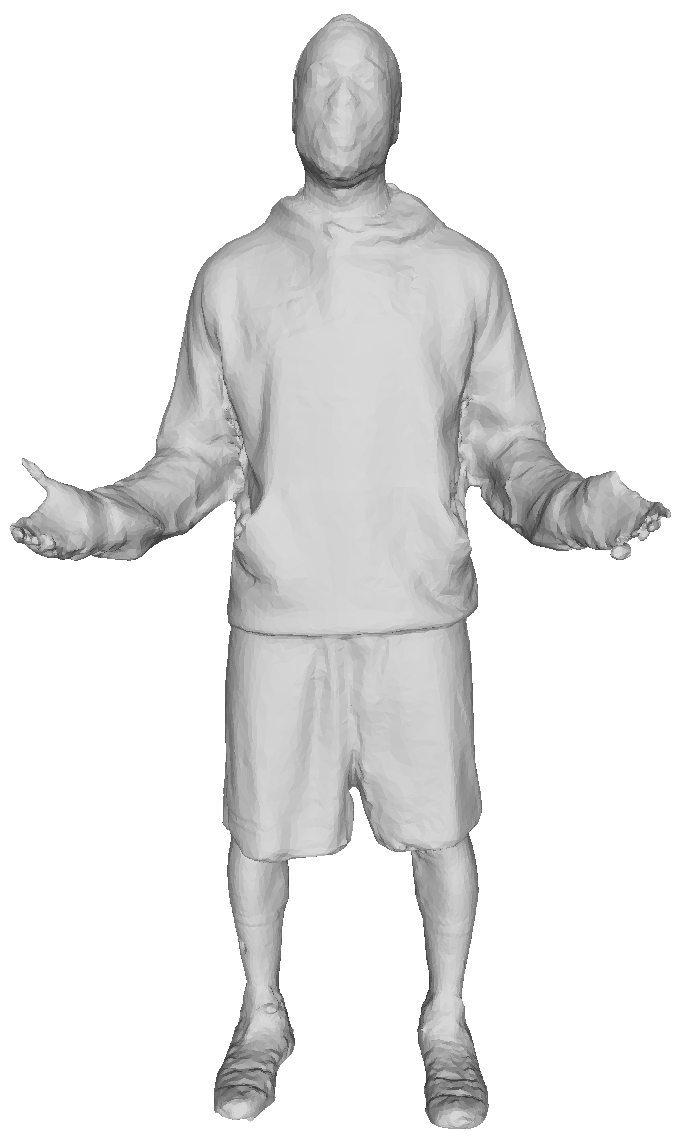}}
\caption{Reconstructed meshes of performing actors.\label{fig_volvideo2}}
\end{figure}

\begin{figure}[htb]
\centering{\includegraphics[ height = 4.5cm ]{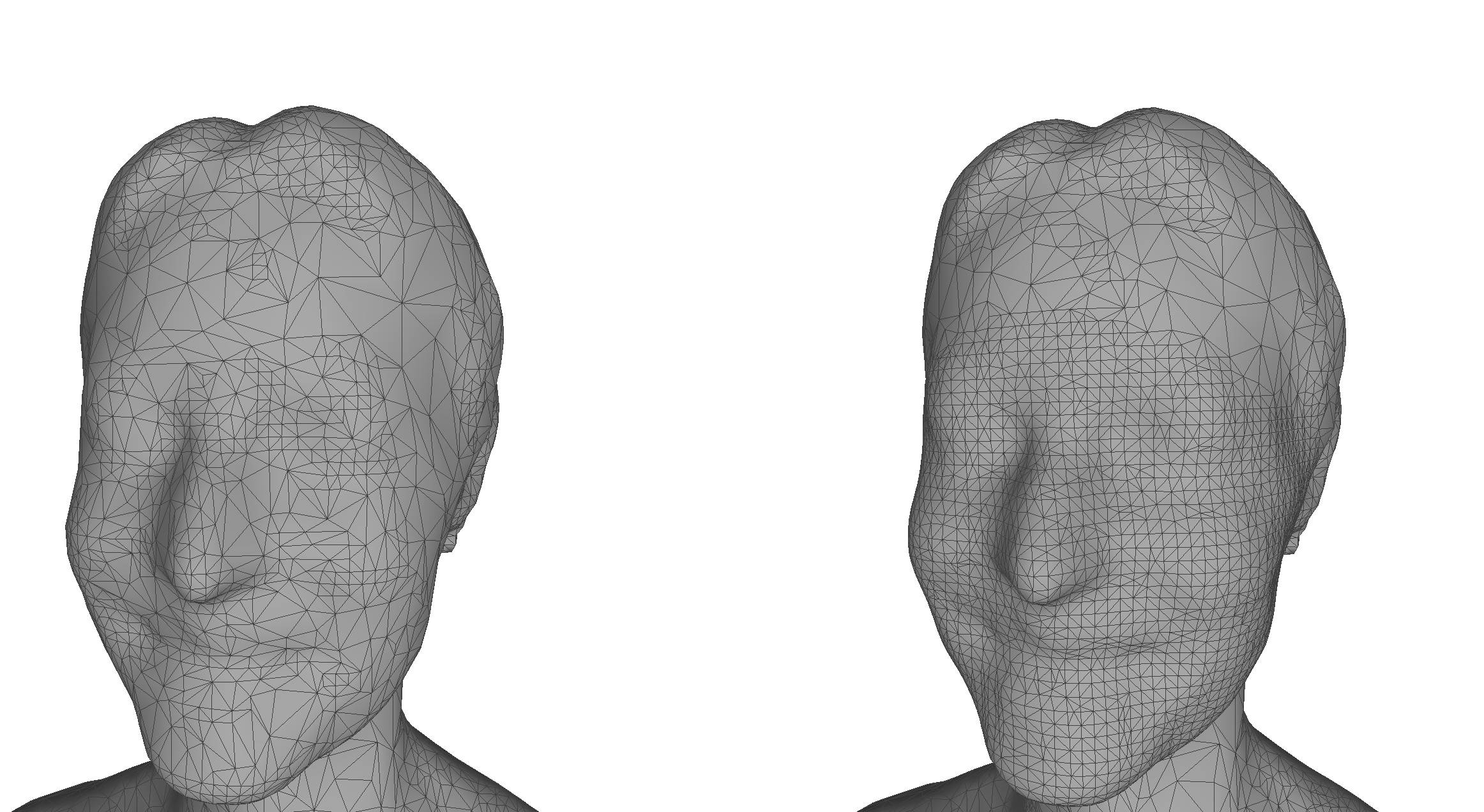}}
\caption{Influence of the face detector on the mesh simplification (left: without face detection, right: with face detection).\label{fig_face}}
\end{figure}

In the next step, the depth maps are fused into a common 3D point cloud \cite{ewrs14}.
The fusion process considers visibility constraints from the individual views as well as normal information taken from the surface patch orientation in order to deal with contradicting surface candidates.
In addition, silhouette information obtained by keying against the bright white background restricts the outer hull.
The resulting point cloud is converted into a triangle mesh using Screened Poisson Surface Reconstruction \cite{kaho13}.
Since this mesh is usually still too large to be rendered on a VR headset or used as an asset in a CG scene, it is further reduced by a modified version of the Quadric Edge Collapse \cite{gahe97}. In contrast to the standard edge collapse approach, we preserve details in semantically important regions like the face (automatically detected by a DNN based face detector combined with skin color classification \cite{laes16}) by assigning a larger number of triangles to these regions \cite{diaz19} (\reffig{fig_face}).
The result of this step is a sequence of 3D meshes as shown in \reffig{fig_volvideo2} with inconsistent mesh connectivity.

\section{Temporal Processing}\label{sec_temporal}

The volumetric pipeline described in the previous section produces a sequence of 3D surface meshes. As frames are processed individually, mesh connectivity may differ substantially between consecutive frames. To facilitate manual editing of textures and corrections of geometry over multiple frames, we aim at enforcing temporally consistent mesh topology. Hence, the next step in our processing pipeline is a mesh registration that provides local temporal stability while preserving the captured geometry and mesh silhouette.

\subsection{Keyframe-based Mesh Tracking for Temporal Consistency}

Establishing the same connectivity for the entire sequence might be infeasible in cases of large scene changes, including surface interaction, object insertion or removal, etc.
Hence, we adapt the keyframe concept from video compression and divide the sequence into groups of frames that will share the same connectivity. For each group, a frame chosen as the keyframe is deformed to match the surface of their neighboring meshes, progressively covering the complete sequence. Frames with a high surface area and low genus are considered to be good candidates to become keyframes in an automatic keyframe selection scheme \cite{collet2015high}. Alternatively, every $n^{th}$ frame can be chosen as the keyframe for sequences, where high input noise might make automatic keyframe selection error-prone.

Once keyframes are chosen, we perform a pairwise mesh registration, deforming the keyframe progressively to its temporal neighbors. We work both temporally forward and backward from the keyframe to reduce the average temporal distance of a registered frame to its keyframe. Following, some frames are selected to be registered from two different keyframes. Greedily selecting the resulting mesh with the smaller registration error from the two candidates improves the smoothness of the transition between keyframes and reduces potential errors from imperfect keyframe selection.

\begin{figure}[htb]
\centering
\includegraphics[width=0.4\textwidth]{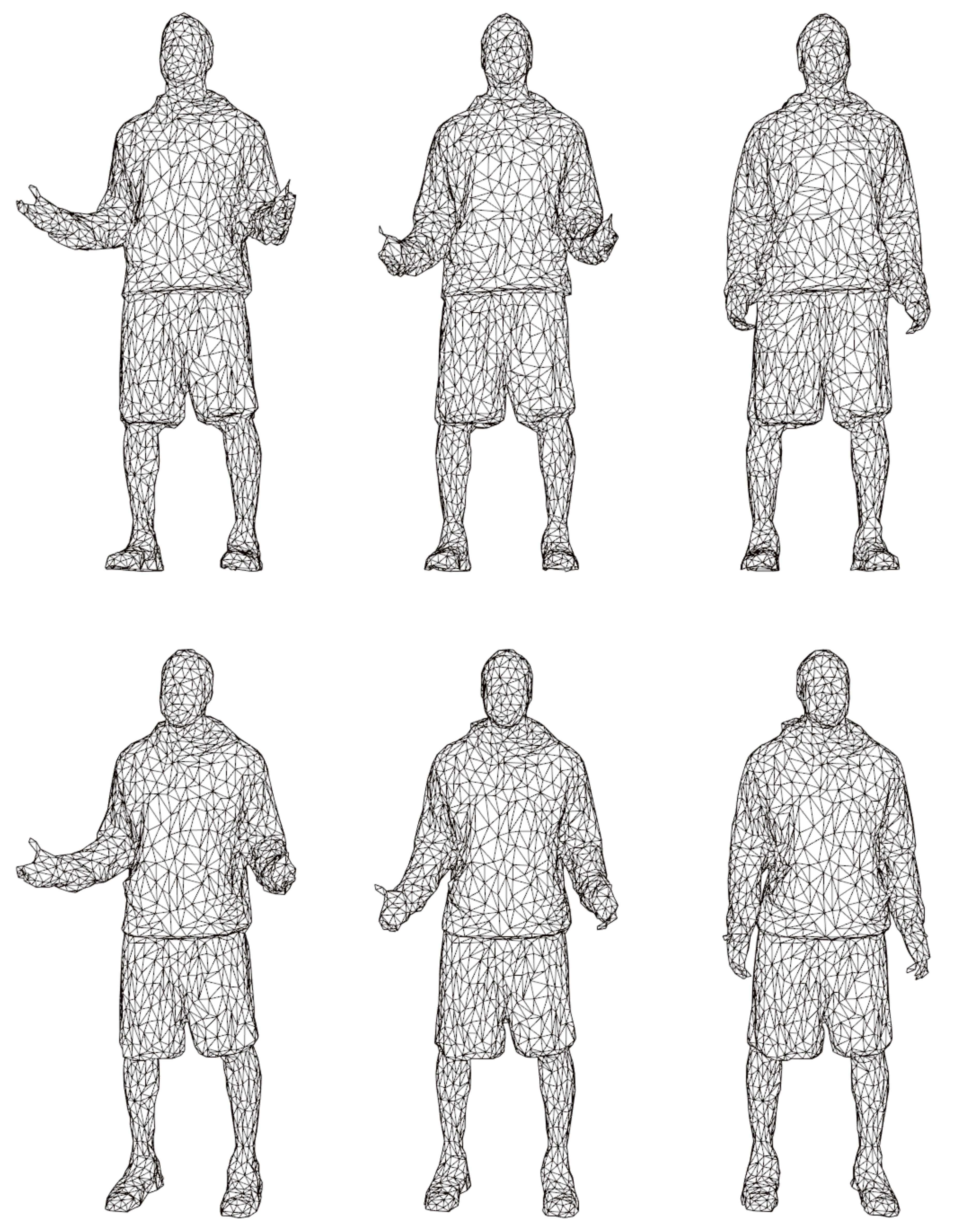}
\caption{Individual meshes of a sequence captured in the volumetric pipeline may have differing connectivity (top row). We perform a mesh registration, deforming the mesh of the first frame into the following frames (bottom row). This provides shared connectivity while preserving the original geometries and silhouettes.}
\label{fig:temporal-consistency-josh}
\end{figure}

Our non-rigid pairwise mesh registration to deform one mesh into another is based on the iterative closest point (ICP) algorithm.
We use a coarse-to-fine scheme to speed up convergence towards a global solution even with large displacements between successive frames. Such a hierarchical ICP scheme reduces computation time while improving robustness \cite{jost2003multi}.

At each level of detail, we run a bidirectional ICP to pull the mesh towards the target surface.
We constrain the deformation to be locally as-rigid-as-possible to preserve local stiffness in articulated objects \cite{sorkine2007rigid}.
The deformation of the mesh is encoded as rotations and translations attached to nodes in a deformation graph \cite{sumner2007embedded}.
Our deformation graph is created from the deformed mesh and does not model any category of objects explicitly.
We extend the idea of the deformation graph to progressively add details to the transformations while working upwards in the coarse-to-fine mesh hierarchy.

With the described non-rigid registration, we create temporally coherent subsequences while preserving the reconstructed geometry.
Through application of the coarse-to-fine approach, we are able to register meshes with $50,000$ faces in a few seconds computation time on a single machine to sub-millimeter accuracy. Details of our temporal processing are described in \cite{mhe19}.

Figure \ref{fig:temporal-consistency-josh} compares three frames of the initial reconstructed mesh sequence with the tracked sequence that is created by deforming the first frame into the following frames.


\begin{figure}[htb]
\centering{\includegraphics[ height = 4.5cm ]{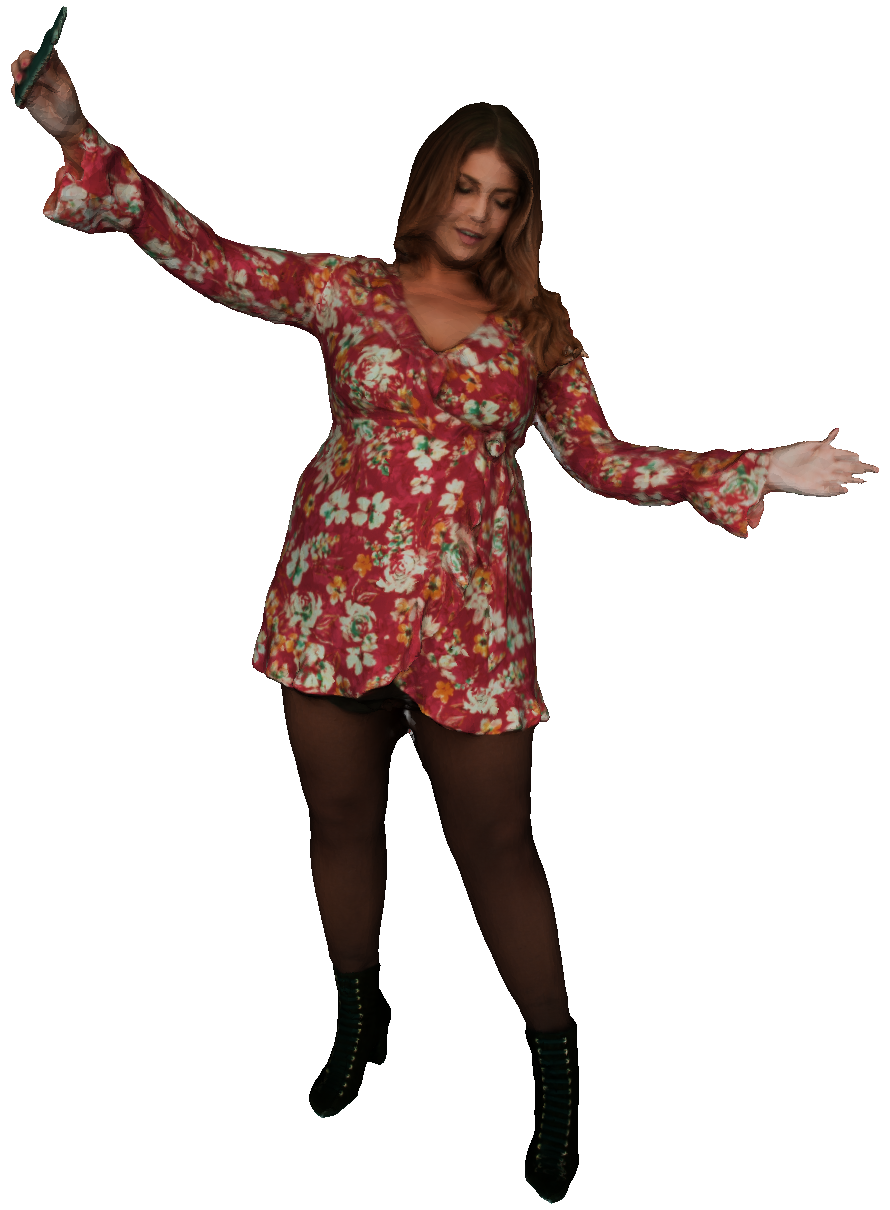}}
\centering{\includegraphics[ height = 4.2cm ]{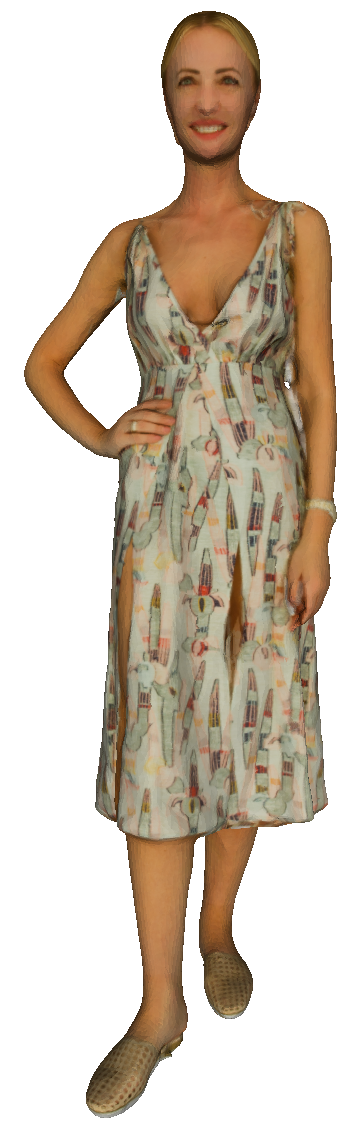}}
\centering{\includegraphics[ height = 4.2cm ]{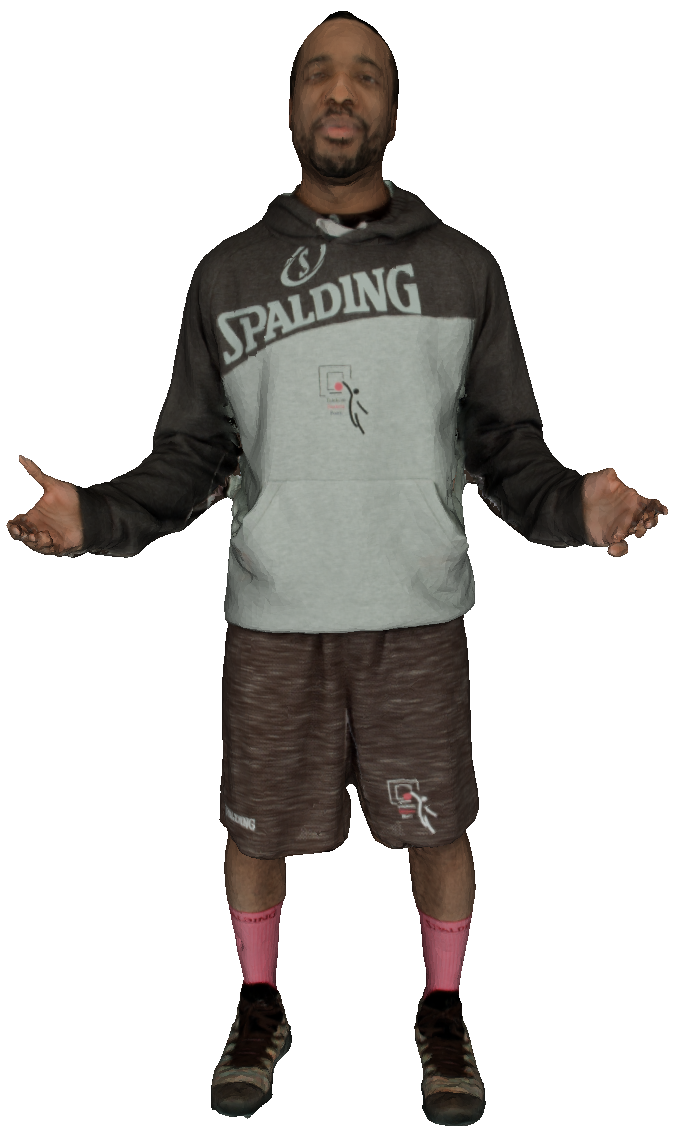}}
\caption{Textured meshes of performing actors.\label{fig_temporal2}}
\end{figure}

\subsection{Mesh Sequence Texturing}

For each subsequence corresponding to a keyframe, we compute texture coordinates and fill the atlas using the captured views.
Since the uv mapping for these frames stays constant, temporal filtering and consistent editing of the texture maps becomes possible.
For texture filling, we extend a graph cut based approach for view selection, color adjustment and filling \cite{wmg14}, incorporating
information on semantically important regions as the face and depth map quality to the data term.
Figure \ref{fig_temporal2} shows examples of such textured models that we have created within several VR productions \cite{sfr19}.
They can be used as assets in virtual environments but provide only free viewpoint visualization without any additional capabilities of animation.

\section{Semantic Enrichment for Pose Editing and Animation}\label{sec_bodymodel}

In the previous sections, we have created high-quality volumetric video.
In order to make the captured data animatable, we fit a parametric kinematically animatable human model enclosing a skeleton to closely resemble the shape of the captured subject as well as the pose of each frame.
Thereby, we enrich the captured data with semantic pose and animation data taken from the parametric model, which can then be used to animate the captured data itself (see Sec.\ref{sec_poseedit}).

\begin{figure}[htb]
  \centering
  \includegraphics[width=0.8\linewidth]{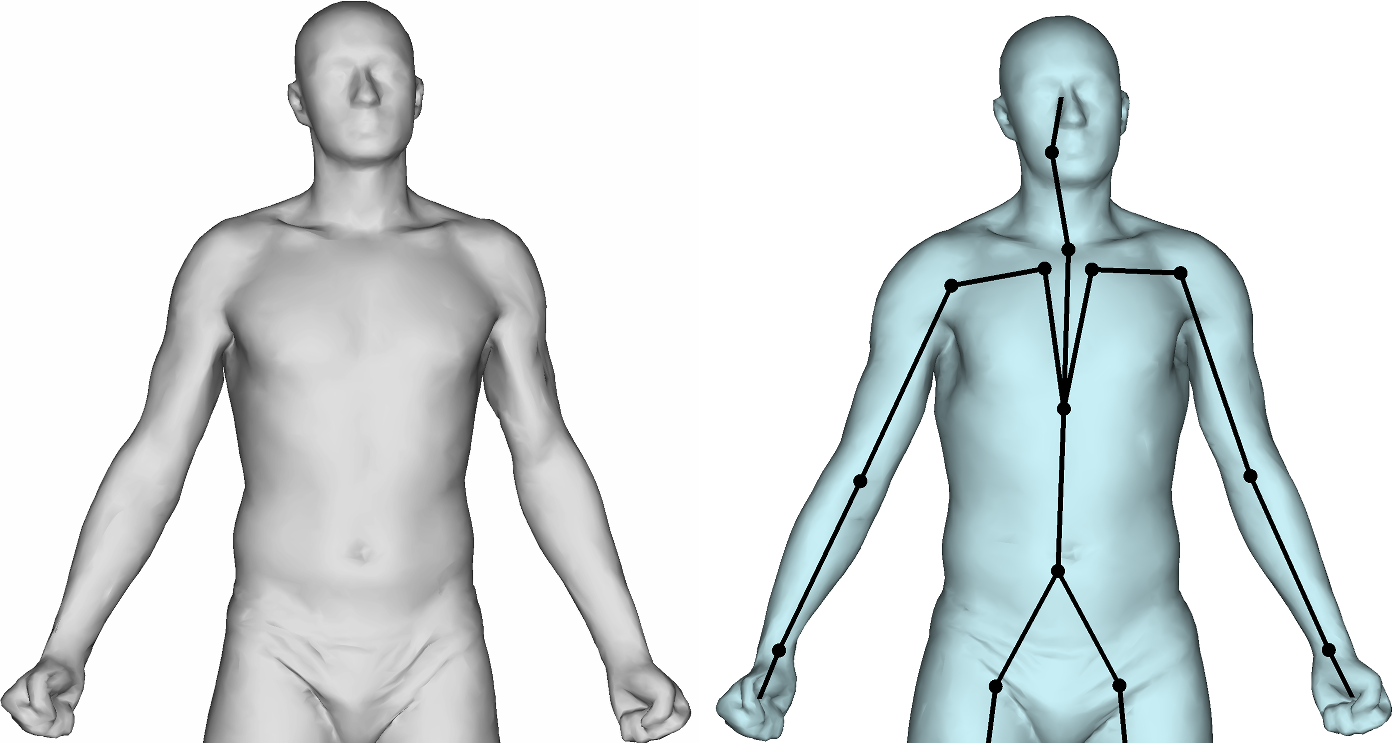}
  \caption{\label{fig:Model} Our articulated model (right) learned from SCAPE dataset (left) \cite{asktrd05}.}
\end{figure}

\subsection{The Body Model}
We want to fit a body model to each frame in order to pass the pose and animation information from the model to the captured data.
In order to produce natural movements and animations, the parametric model's shape should resemble the captured data as closely as possible.
We follow the approach of \cite{fhe16} and learn our shape-adaptive human model (skinning weights, vertices, kinematic joint position and orientation) in a data-driven optimization from datasets of captured humans resulting in a high degree of accuracy and natural realism (\reffig{fig:Model}).
We base our model on vertex skinning in order to allow real-time animation, in contrast to implicit animation methods \cite{Hasler:EUROGRAPHICS2009, SM:CVPR:2015, Bogo:ICCV:2015, Dyna:SIGGRAPH:2015, Meekyoung:siggraph2017}.
In order to achieve natural and realistic movements,
our deformation model is based on a swing-twist decomposition for joint rotations \cite{fhe16}.
The two main advantages of this parameterization are:
  (i) The parameters are interpretable. Thus, it is straightforward to specify lower/upper bounds for increased robustness of tracking algorithms, or to extract these bounds automatically from example datasets \cite{mbhks19}.
  (ii) Animating the model with different skinning functions,
      i.e. Linear Blend Skinning for swing rotations and Quaternion Blend Skinning for twist,
      reduces skinning artifacts to a non-visible minimum \cite{ks12},
      especially when using the dual quaternion based multi-joint variant \cite{fhe16},
      without requiring further enhancements like blend shapes.

\subsection{Body Model Fitting}
\begin{figure}[htb]
  \centering
  \includegraphics[width=1.0\linewidth]{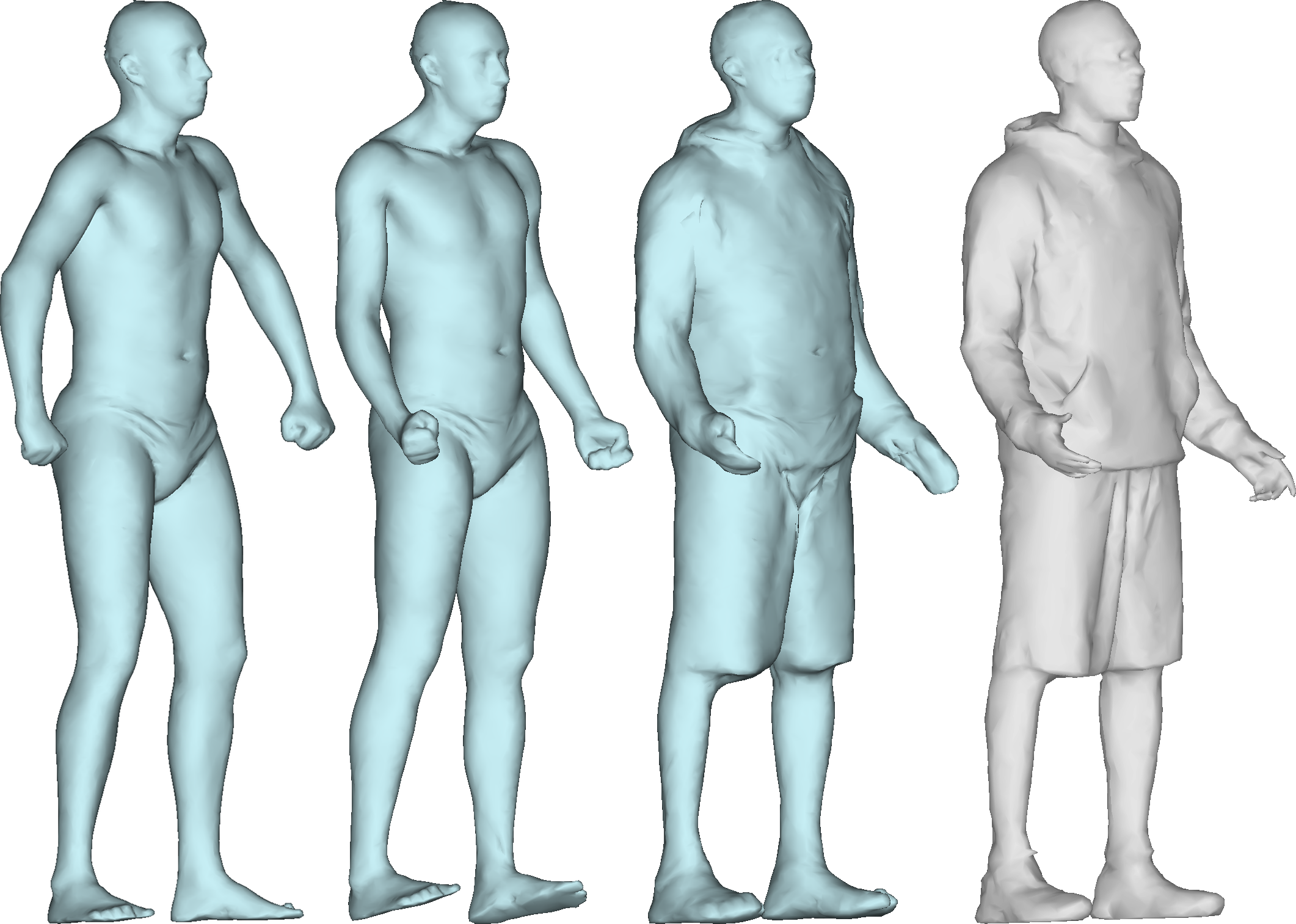}
  \caption{\label{fig:ModelFitting} Model adaptation to volumetric video data: initial model (left), pose adapted model (2$^\textrm{nd}$ from left), pose and shape adapted model (2$^\textrm{nd}$ from right) and volumetric video frame (right).}
\end{figure}

We fit the articulated model to the sequence of 3D meshes using shape adaptive motion capture as presented in \cite{fhe19}.
Starting from a rough manual kinematic alignment for the first frame,
we optimize the pose parameters by minimizing the distances between corresponding vertices of the model and the captured mesh.
This initial pose fit is refined in a second step by additionally optimizing the model vertices and skeleton joints to bring the model into better alignment with the captured mesh
as shown in \reffig{fig:ModelFitting}.
The resulting subject-adapted model is used to track the persons pose from frame to frame.
To ensure natural movement characteristics, we use a learned pose prior based on Gaussian mixture models.
Further, we employ a mesh Laplacian constraint \cite{LaplacianSTAR:2006} to enforce plausible human shapes.
Details on the fitting of the body model can be found in \cite{fhe19}.

Through fitting the kinematic body model to the individual frames of the volumetric video data, we enrich the captured real data with semantic information on pose and animation properties, making the volumetric video data animatable.

\section{Pose Editing and Animation}\label{sec_poseedit}

\begin{figure*}[t]
  \centering
  \includegraphics[width=1.0\linewidth]{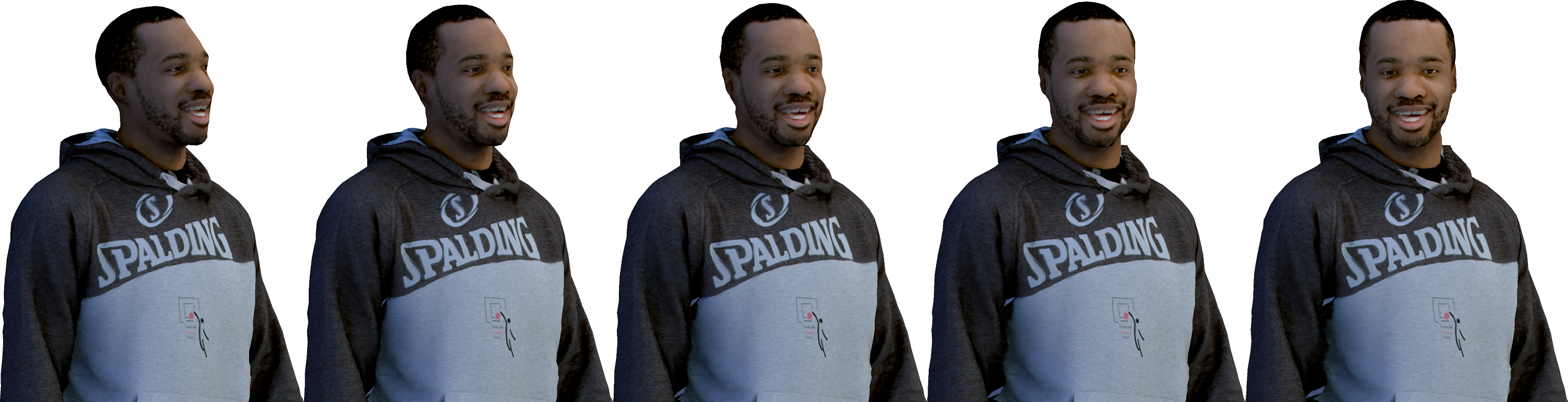}
  \caption{\label{fig:JoshScanAnimated}Captured/original frame of the volumetric video data (1$^\text{st}$ image on the left) and animated (gaze corrected) volumetric video frames (all remaining images on the right).}
\end{figure*}

For the generation of new performances with real deformations (present in the original data), we propose to follow a hybrid example-based approach that exploits the captured data as much as possible. As the captured volumetric video data consists of temporally consistent subsequences, these can be treated as essential basis sequences containing motion and appearance examples. The enrichment of the volumetric video data with pose semantics allows us to retrieve the subsequence(s) or frames closest to a target pose sequence, which are then concatenated and interpolated in order to form new animations similar to surface motion graphs as described in \cite{bb15, bb19}. For smoothing the transitions between successive sequences, the progressive mesh tracking approach described in Sec.~\ref{sec_temporal} can be used in order to register the meshes of one sequence to the other, and then interpolating over time from the original sequence to the registered sequence.

The new sequences generated by this approach are restricted to poses and movements already present in the captured data and might not perfectly fit the desired poses.
However,
as we enriched the volumetric video data with animation and pose properties in the previous section,
we can now kinematically adapt the recomposed frames to fit the desired pose (see below).
Put another way,
the individual volumetric frames are animatable but in order to attain a high realism,
we want to change each frame as little as possible.
Hence,
we first generate a close animation sequence in an example-based manner as described above and only slightly modify the recomposed frames to fit the target poses.
This hybrid strategy combines the realism of the real volumetric video data with the ability to animate and modify.


The kinematic animation of the individual frames is facilitated through the body model fitted to each frame.
For each mesh vertex, the location relative to the closest triangle of the template model is calculated, parameterized by the barycentric coordinates of the projection onto the triangle and the orthogonal distance. This relative location is held fixed, virtually glueing the mesh vertex to the template triangle with constant
distance and orientation.
With this parametrization between each mesh frame and the model, an animation of the model can directly be transferred to the volumetric video frame.
Fig.~\ref{fig:JoshScanAnimated} shows an example of an animated volumetric video frame.
The classical approach would have been to use the body model fitted to best represent the captured data in order to synthesize new animation sequences. Instead, exploit the captured data as much as possible and use the pose optimized and shape adapted template model to drive the kinematic animation of the volumetric video data itself, exploiting the fact that the captured data exhibits all details and natural movements. As the original data contains natural movements with all fine details and the original data is exploited as much as possible, our animation approach produces highly realistic results. Of course, the variety of deformations and movements that can be synthesized solely from the original data is limited by the variety of movements that have been captured. Although we keep the animation and deformation of the captured data as minimal as possible, especially loose clothing will benefit from an additional or more sophisticated underlying animation method than a statistical body model, e.g. based on works that model highly non-rigid deformation of clothing as a function of body pose \cite{hfe13,soc19, gundogdu18}.



\section{Hybrid Facial Animation}\label{sec_face}

While purely geometry-based animation approaches are very popular and work well for large scale deformation and human pose animation, they are usually too limited in expressiveness for realistic facial animation and rendering. Especially the mouth and eye area exhibit strong non-linear changes in geometry and appearance, which are difficult to represent with blendshapes or skeleton joints. Alternative animation approaches use image data to directly synthesize new facial expressions from previously captured images. Such image-based animation techniques provide usually high-quality renderings but they are not as flexible as computer graphic models. We propose a new facial animation method combining the advantages of both strategies: the flexibility of computer graphic models with the realism of image-based rendering.

\subsection{Video-Based Facial Animation}

\begin{figure}[htb]
\centering
\includegraphics[width=0.7\columnwidth]{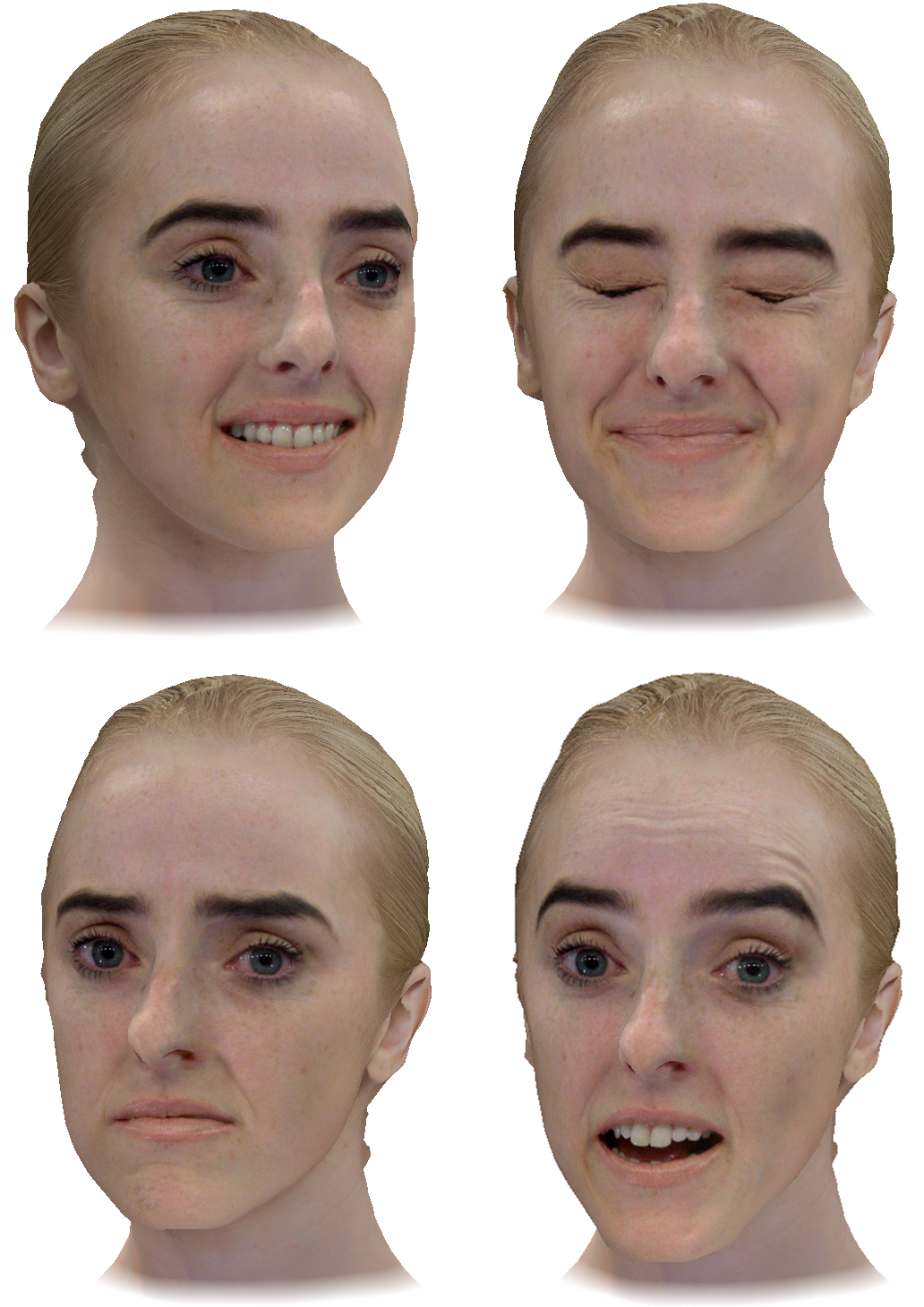}
\protect\caption{Four examples of our facial animation model, rendered with different facial expressions and different viewpoints. }
\label{fig:oldresults}
\end{figure}

Our facial animation models consist of two parts: we use the captured face geometry to explain rigid motion and large scale deformation and add a dynamic texture model that represents all details, small movements and changes in appearance (e.g. small wrinkles or local variations of skin colour due to strong deformations, see \reffig{fig:oldresults} (forehead and around mouth).

From the captured volumetric video data, we compute a set of personalized blendshapes with a consistent topology (face geometry proxy). In \cite{pkhe15}, we show that it is even sufficient to have a face proxy with only one degree of freedom for deformation jaw) plus six degrees of freedom for rigid motion. We intentionally do not remove mouth and eyes from the computed blendshapes, since we need a complete face geometry to render our dynamic textures.

The second part of our animation strategy is constituted by dynamic face textures, which are extracted from the volumetric video data using a graph-cut based approach \cite{pkhe15}. We use the previously created face geometry proxy to estimate the 3D face pose (i.e. 6 DoF + deformation) in every captured video frame and extract a temporal consistent sequence of face textures \cite{pkhe17,pkhe15}.

\begin{figure}[htb]

\includegraphics[width=\columnwidth]{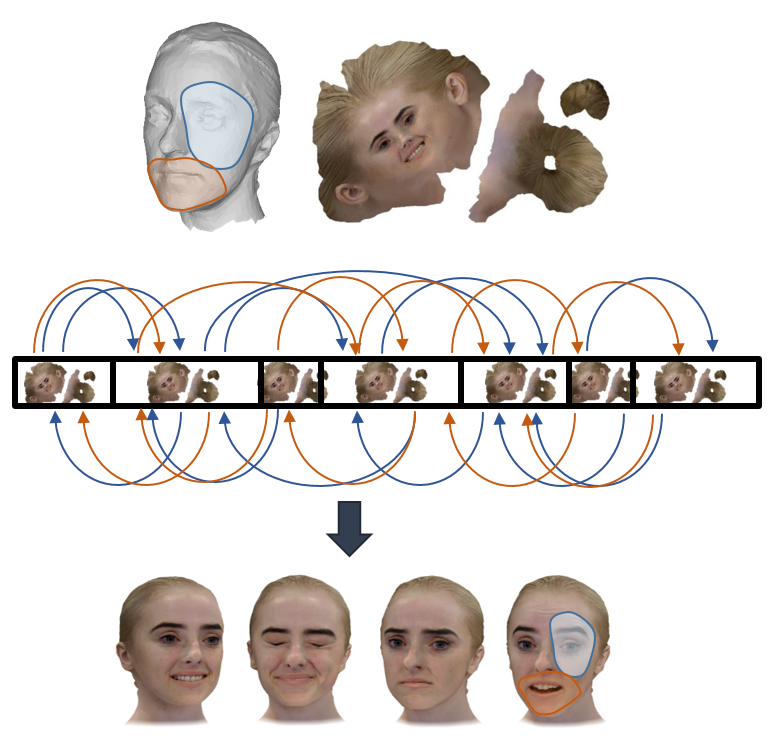}
\protect\caption{Short sequences of the volumetric video footage are re-arrange, looped and concatenated to form a new facial animation sequence. We can even subdivide the dynamic textures into local regions (e.g. eye region in blue, mouth region in orange) in order to reassemble the animation from different captured sequences.}
\label{fig:texanim}
\end{figure}

The actual animation methodology is closely related to motion graphs \cite{Kovar2002,Casas2012}. We use short sequences of the captured video footage and re-arrange, loop and concatenate them to create novel video sequences, see \reffig{fig:texanim}.
In the context of motion graphs, edges in the graph correspond to facial actions, and vertices to expression states. Since we have approximate 3D information, we are able to compensate for the global head pose, which allows us transferring facial expressions even between videos with different head orientations \cite{pkhe17}.
Since the extracted sequences have been captured separately and in a different order, simple concatenation would create obvious visual artifacts during transitions between two sequences.
These artifacts appear in geometry as well as in texture due to different facial expressions and changing illumination.
The independent texture sequences are brought into connection by defining transition points between the separate sequences \cite{pkhe17}, see \reffig{fig:texanim}.
We can also subdivide the texture sequences into subregions, e.g.~the eyes and mouth and define transition points for each region separately, such that the different facial regions of the new animation sequence can be assembled from different source sequences, broadening the range of possible animations.

In order to create seamless transitions between concatenated texture sequences, we use a spatio-temporal blending approach that explains texture mismatch with optical flow using a mesh-based motion and deformation model \cite{hilsmann2008}.
Using the optical flow information, we can smoothly interpolate between different facial expression textures without creating ghosting artifacts.
All texture-differences that cannot be compensated by optical flow alone, are blended with a cross dissolve approach.

\subsection{Neural Infilling}
While the spatio-temporal blending works well is in most cases it can fail if two concatenated facial expressions are too different.
For example, disocclusions or occlusions caused by an opened or closed mouth cannot be explained by optical flow. Therefore, we make use of recent advances in deep convolutional neural networks that are capable of learning generative models for images and textures.

\begin{figure}[htb]
\includegraphics[width=\columnwidth]{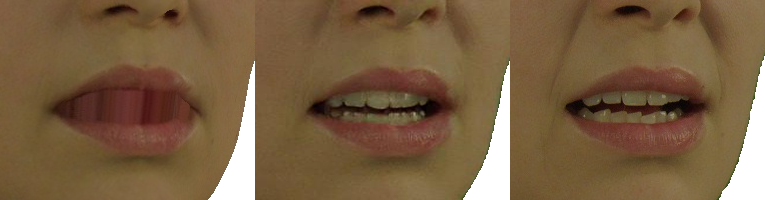}
\protect\caption{Left: Example of a rendered image with missing oral cavity. Middle: corrected image with synthesized oral cavity by a cGAN. Right: Ground truth}
\label{fig:cgan}
\end{figure}

In order to address these problems, we implemented two methods based on deep neural networks. As a first approach, for example, we train a conditional GAN to correct artifacts that occur during animation.
The main idea of this approach is to interpret artifacts occurring during animation as features. For this purpose, we train an image-to-image translation network that is capable of inferring a realistic mouth image from a rendered mouth showing animation artifacts.
For this experiment, we use a U-Net like architecture \cite{Ronneberger2015} as an image generator and a PatchGAN as discriminator like in \cite{Li2016}. The generator's task is to infer the realistic mouth picture from a rendered mouth image that contains artifacts.
The discriminator receives image pairs that contain the rendered version plus the corrected/captured mouth image. Base on these image pairs, it has to decide if the artifact-free image is real or if it was synthesized by the generator network.
In \reffig{fig:cgan}, we show an animated face with an open mouth. Due to the deformation, visual artifacts appear since the oral cavity is not represented in texture.
Using our GAN that is conditioned on visual artifacts that appear during animation, we can reconstruct the expected artifact-free image (e.g. with visible teeth).

\begin{figure}[htb]
\includegraphics[width=\columnwidth]{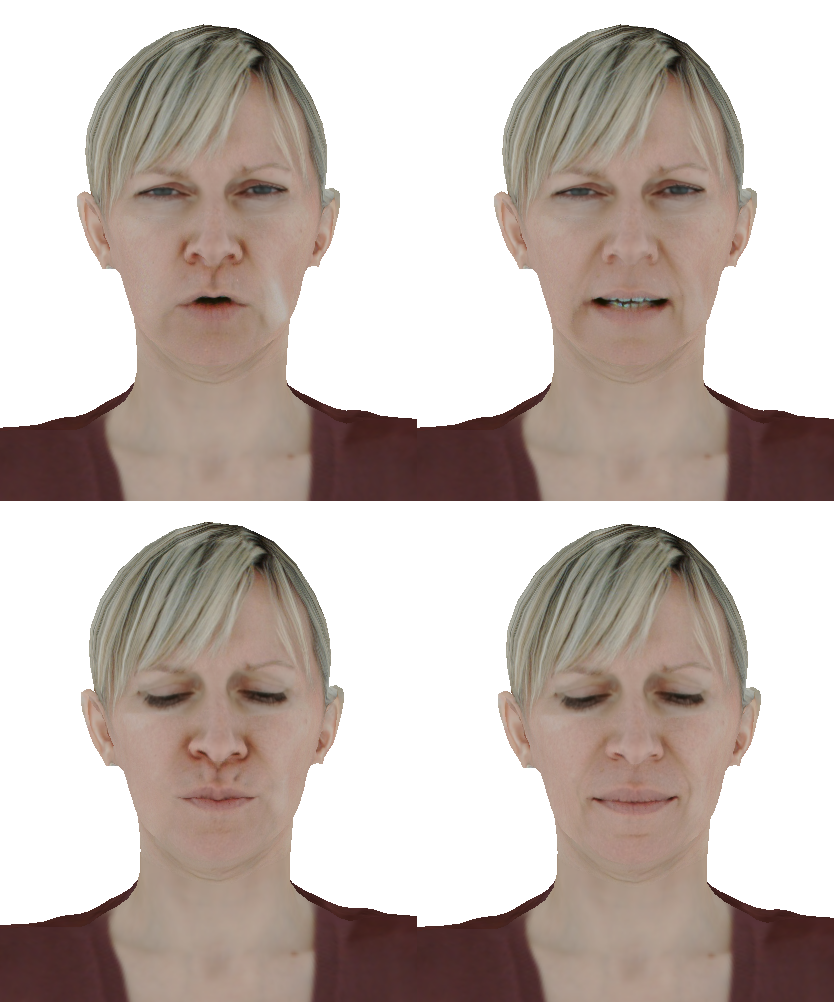}
\protect\caption{Examples of the neural face model with dynamic textures synthesized by a variational autoencoder. We use two local models for eyes and mouth in order to animate them independently from each other.}
\label{fig:vae}
\end{figure}

A second approach for the improvement of our animation pipeline is based on variational autoencoders \cite{Kingma2014}. The main idea behind this approach is, to directly avoid animation artifacts by learning a complex generative model that can capture the high dimensional space of texture and geometry
and is interpolates between different samples without introducing blending artifacts. Similar to \cite{Lombardi2018}, we train local variational autoencoders that capture geometry and texture of the mouth and eye region. Each local autoencoder provides a nonlinear mapping from a high dimensional raw-data space (i.e. textures and vertex coordinates)
to a low dimensional latent space and back. The latent space can be interpreted as a compressed version of raw mesh data with additional constraints such that there is a smooth transition between facial expressions and similar facial expressions are mapped close to each other. These constraints allow smooth interpolation between facial expressions.
Our overall animation strategy stays the same, but instead of working with raw data directly we manipulate facial expressions in latent space. This allows us to blend smoothly and free of artifacts between concatenated sequences, see \reffig{fig:vae} for examples of synthesized facial expressions.
Another advantage of the variational autoencoder is that raw image and geometry data can be compressed into a low dimensional latent representation, which also reduces memory requirements on the rendering machine.

\section{Conclusion}\label{sec_conclusion}
In this paper, we have presented a full end-to-end pipeline for the creation of high-quality animatable volumetric video of human performances.
The idea is to enrich the captured data with semantic pose and animation properties in order to allow a modification of the individual volumetric video frames.
To ensure a maximum of realism of the synthetic sequences, we exploit the captured data, exhibiting all the fine real deformations and appearance changes during motion, as much as possible in a cascading example-based approach:
Temporal consistency allows a re-composition of existing frames or subsequences in order to fit a desired animation sequence as close as possible, followed by a kinematic mesh modification.
As humans are very sensitive to seeing facial expressions, we treat the face separately from the body and propose a hybrid geometry- and video-based approach, which uses a coarse geometric model for large-scale deformation and global motion and video-based textures to represent subtle and detailed facial movements.
Finally, missing regions are realistically filled in using a neural texture synthesis approach.
The full hybrid animation framework combines the realism of high-quality volumetric video with the flexibility of Computer Graphics methods for animation.

Our pipeline and the hybrid animation approach are designed to facilitate the integration of alternative modules e.g. during capturing or additional or more sophisticated underlying animation approaches, e.g. approaches that can also model the deformation of clothing or other physics. Of course, the variety of deformations and movements that can be synthesized solely from the original data is limited by the variety of movements present in the data. We keep the animation and deformation of the captured data as minimal as possible. However, the current underlying animation approach might not be able to fully represent the deformation of loose clothing if this is not present in the data. Here, current works on learning pose-dependent highly non-rigid deformation of clothing as a function of body pose \cite{hfe13,soc19, gundogdu18} or data-driven soft tissue animation \cite{Meekyoung:siggraph2017, CasasO18} will add value to the presented hybrid approach.

\section{Acknowledgments}\label{sec_acknowledgement}

This work has partly been funded by 
the European Union’s Horizon 2020 research and innovation programme under grant agreement No 762021 (Content4All) 
as well as the Germany Federal Ministry of Education and Research under grant agreement 03ZZ0468 (3DGIM2).
We thank the Volucap GmbH for providing the Josh sequence that was recorded for the MagentaVR app.

\bibliographystyle{IEEE}
\bibliography{ietcv2019, ietcv2019_face}

\end{document}